\documentclass[11pt]{article}
\usepackage[final]{acl}
\usepackage{times}
\usepackage{latexsym}
\usepackage[T1]{fontenc}
\usepackage[utf8]{inputenc}
\usepackage{microtype}
\usepackage{inconsolata}
\usepackage{graphicx}
\usepackage{booktabs}
\usepackage{amsmath}

\title{The Attribution--Compression Frontier in Retrieval-Augmented Generation}

\author{Deepanshu Mody \\
  Center for Data Science \\
  New York University \\
  \texttt{dm6262@nyu.edu}}

\hypersetup{
  pdfauthor={Deepanshu Mody},
  pdftitle={The Attribution-Compression Frontier in Retrieval-Augmented Generation}
}

\begin{document}
\maketitle

\begin{abstract}
Context compression reduces generator input in retrieval-augmented generation, but answer quality alone does not characterize citation attribution.
We measure citation attribution across compression methods and budgets, comparing reranking, extractive selection, abstractive summarization, token pruning, and an extract-cluster-rewrite construction on ASQA and QASPER under a fixed generator and primary entailment evaluator.
On ASQA at a nominal 0.25 budget (achieved compression 0.08), a RECOMP-style compressor's citations score 0.86 precision against its summaries but 0.12 against source spans under our re-attributability protocol.
These estimates depend on a shared NLI model for span recovery and citation scoring and lack independent human calibration.
Extractive selection's observed grounded precision ranges from 0.43 to 0.49 across nominal budgets from one-half to one-tenth of the ASQA context, while answer quality declines.
For the same RECOMP setting, claim verification after source recovery yields an unsupported rate of 0.88 versus 0.17 when checking summaries.
This gap persists beyond structural rejection of missing provenance, but remains evaluator-dependent.
A 200-question TRUE T5-XXL audit also finds emitted--grounded gaps under both fixed and recomputed source mappings, without establishing human-calibrated support rates.

\end{abstract}

\section{Introduction}

Context compression reduces generator input in retrieval-augmented generation (RAG) \citep{xu2024recomp,jiang2024longllmlingua,pan2024llmlingua2,chirkova2025provence}.
Citation-grade attribution requires \emph{locatable} evidence: a reader must be able to retrieve and check the cited source \citep{rashkin2023ais,bohnet2022attributed}.
Abstractive compression can interrupt this link when citations identify compressor-generated text without its supporting source spans.
We call it \emph{attribution laundering} when consistency with compressor-generated text is treated as evidence of support from original sources.
On ASQA at nominal budget $0.25$ (achieved ratio $0.08$), a RECOMP-style compressor's citations score 0.86 precision against its summaries and 0.12 against source spans under our recovery protocol.
These automatic scores depend on a shared NLI evaluator and bounded candidate search; they have not been independently calibrated.

Related work already identifies a divergence between answer quality and citation grounding under compression \citep{li2026verifiability}, while citation support does not establish causal reliance on evidence \citep{wallat2025faithfulness}.
We examine this problem across compressor families and two datasets, reporting achieved context fractions, explicit source-recovery rules, and paired comparisons on shared questions.
Our measurements concern source support under an automatic evaluation protocol; they do not test causal citation faithfulness.

We conduct a controlled measurement study with a fixed generator (Qwen2.5-7B-Instruct with ALCE-style numbered-context citation prompting), a fixed primary NLI evaluator, and nominal budgets of $\{0.5, 0.25, 0.1, 0.05\}$ in generator-tokenizer tokens.
Our contributions are:

\begin{itemize}
  \item \textbf{The attribution--compression frontier.} A comparison across reranking, extractive selection, abstraction, token pruning, ECR, and an identity reference against achieved context fractions.
  All settings share 944 successful ASQA questions and 976 QASPER questions from 271 papers; paired tests quantify their differences.
  \item \textbf{A re-attributability protocol.} A unit without declared provenance is counted as re-attributable when NLI predicts entailment from a candidate source snippet; recovered sources support \emph{grounded} citation scores alongside \emph{emitted} scores against compressed text.
  \item \textbf{Three verification modes.} We separate consistency with summaries, availability of declared provenance, and claim support after source recovery.
  Structural rejection under declared-only checking is distinguished from measured rejection after recovery.
  \item \textbf{An extract-cluster-rewrite (ECR) observation.} A simple construction greedily merges near-duplicate sentences and cites \emph{all} members of each cluster; paired comparisons quantify its measured differences from extraction across budgets.
\end{itemize}

\section{Related Work}
\label{sec:related}

\paragraph{Context compression for RAG.}
Text compressors select or rewrite retrieved material.
Selection approaches include query-aware token pruning \citep{jiang2024longllmlingua}, distilled token classification \citep{pan2024llmlingua2}, sentence pruning with reranking \citep{chirkova2025provence}, and evidentiality-guided selection \citep{jeong2025ecorag}.
RECOMP trains extractive and abstractive compressors for end-task performance \citep{xu2024recomp}, while FaviComp combines retrieved context with parametric knowledge during rewriting \citep{jung2025favicomp}.
Whether their outputs retain locatable source links depends on the integration: our extractive and ECR implementations declare source IDs, while our LLMLingua-2 integration omits a token-to-source mapping.
Soft compressors use learned representations, including gist tokens \citep{mu2023gist}, ICAE memory slots \citep{ge2024icae}, document embeddings \citep{cheng2024xrag}, and multi-context embeddings \citep{rau2025cocom}; these are outside our experiments.

\paragraph{Attributed generation and citation evaluation.}
The AIS framework evaluates whether statements are supported by identified sources \citep{rashkin2023ais}.
Automatic attribution and entailment evaluation \citep{bohnet2022attributed,honovich2022true} inform the citation precision and recall metrics of ALCE \citep{gao2023alce}, which we adapt.
QASPER provides questions over research papers with annotator-marked evidence \citep{dasigi2021qasper}.
LongCite studies fine-grained citation generation with long contexts \citep{zhang2025longcite}, a complementary direction to evaluating compression under a fixed generator.

\paragraph{Compression and citation reliability.}
\citet{li2026verifiability} study Self-RAG on ASQA with LLMLingua-2 and prefix truncation, finding sharper degradation in citation grounding than answer quality, and investigate hierarchical evidence retention.
Our study adds cross-family comparisons on ASQA and QASPER, distinguishes declared and recovered source mappings, and reports emitted and grounded scores against achieved context fractions.
The conceptual finding overlaps; our contribution is experimental coverage and explicit evaluation and reporting procedures, rather than priority for the divergence.
\citet{wallat2025faithfulness} distinguish citation correctness from causal faithfulness and investigate post-rationalization.
Our NLI-based support scores do not establish whether the generator causally relied on the cited evidence.
ALCE's summary-context ablation also illustrates a correctness--citation trade-off \citep{gao2023alce}; information-preservation studies examine grounding under compression \citep{lajewska2025info}, and human evaluations examine verifiability as output abstractiveness changes \citep{worledge2024spectrum}.

\paragraph{Provenance and structured representations.}
Related provenance-oriented designs include claim-anchored multi-document summarization in CAMS \citep{guan2026cams}, budgeted evidence retention in agent memory \citep{li2026ember}, and provenance-aware tiered memory \citep{zhu2026tiermem}.
Structured retrieval indexes use atomic propositions \citep{chen2024densex}, recursive summary trees \citep{sarthi2024raptor}, or graph community summaries \citep{edge2024graphrag}.
These designs motivate explicit links between condensed representations and source evidence, but are not evaluated as compressors here.
Attention-based explanations remain contested \citep{jain2019attention,wiegreffe2019attention}.
Slot-based representations \citep{locatello2020slot,tan2025smarte,ibtehaz2026proteinstory} suggest another possible compression mechanism; whether their internal structure can provide validated source attribution is an untested future direction.

\section{Measuring Attribution Under Compression}
\label{sec:protocol}

\subsection{Setup}
\label{sec:protocol-setup}

Our pipeline is retrieve $\rightarrow$ compress $\rightarrow$ generate $\rightarrow$ evaluate.
A compressor maps the retrieved snippets (sentence-segmented, each with a stable source id, \emph{sid}) to a sequence of \emph{units}, and every unit either declares provenance---the set of sids it was selected from, as in extractive families---or declares none, as in abstractive families.
The generator sees the units as a numbered context and is prompted, ALCE-style, to attach bracketed unit citations to each answer sentence \citep{gao2023alce}.
Citation quality is scored per answer statement with an NLI evaluator in the attributable-to-identified-sources sense \citep{rashkin2023ais,bohnet2022attributed}: a statement is \emph{recalled} iff the concatenation of its resolvable cited sources entails it, and a citation is \emph{precise} iff that concatenation entails the statement and, in addition, the cited source alone entails it or removing that source breaks the concatenation's entailment \citep{gao2023alce}.
We use DeBERTa-v3-base (MNLI+FEVER+ANLI) at threshold $\tau{=}0.5$, truncating each NLI input to 512 model tokens.
Entailment decisions below are this model's predictions, not independently validated judgments of source support.
Per-example harness failures occur only in identity (4 of 948 on ASQA, 29 of 1{,}005 on QASPER); every primary setting excludes those same question IDs, with failure counts retained.

\subsection{Emitted vs.\ Grounded Citations}
\label{sec:protocol-grounded}

We score every run under two views.
We write precision as $P$ and recall as $R$, with subscripts $e$ for emitted and $g$ for grounded.
\textbf{Emitted} scores citations against the unit texts the generator actually saw; it measures consistency with the compressed context, and is what one implicitly measures when compressed units are treated as citable documents.
\textbf{Grounded} first resolves each citation to retrieved source spans and scores against those.
Units with declared provenance resolve to their declared sids.
Units without declared provenance pass through our \emph{re-attributability} protocol: rank source snippets by lexical similarity to the unit, run NLI on the top 10 candidates, and keep those judged to individually entail the unit at $\tau$.
A citation with no recovered span receives no precision credit and cannot support recall under this protocol.
The \emph{re-attribution rate} is the fraction of units without declared provenance with at least one recovered span, averaged across examples.

\paragraph{Evaluator dependence.}
The same NLI model filters recovered spans and scores citation support, creating a circularity risk: selection and evaluation errors can be correlated.
Lexical retrieval can miss paraphrases, and the single-snippet test can reject units supported only by combined evidence.
Thus failed recovery does not establish that support is absent from the source, and the direction of overall scoring bias is unknown.
Appendix~\ref{app:true-audit} reports a second-evaluator sensitivity check; human calibration remains absent.

The emitted$-$grounded gap is an operational diagnostic of \emph{attribution laundering}: claims receive credit against compressed text that is not supported through the source mappings accepted by our protocol.
It does not by itself distinguish missing evidence from recovery or NLI errors.

\paragraph{Nominal vs.\ achieved compression.}
Budgets are nominal fractions $\{0.5, 0.25, 0.1, 0.05\}$ of each example's retrieved context, enforced in generator-tokenizer tokens.
Extractive families track their budgets closely, but abstractive compressors emit what their decoder emits: RECOMP realizes achieved ratios of $0.05$--$0.08$ on ASQA and $\approx 0.015$ on QASPER across all four nominal budgets---distinct nominal operating points collapse together.
All curves therefore use the \emph{achieved} ratio on the compression axis; plotting against nominal budgets would misplace abstractive systems by more than an order of magnitude in the worst case.

\subsection{Three Verification Modes}
\label{sec:protocol-verify}

We decompose each answer into claims (its statements with citation markers stripped; cf.\ atomic-fact decomposition, \citealp{min2023factscore}) and check each claim against evidence pooled from its cited units.
\textbf{Summary-checked} ($U_m$) pools the cited units' own texts.
\textbf{Span-checked} ($U_s$) pools only the \emph{declared} source spans of the cited units; claims without such evidence count as unsupported without an NLI call.
We retain the label ``span-checked'' for this declared-only mode, which differs from grounded citation scoring: it does not recover missing spans.
Its unsupported rate of $1.00$ for methods without declared provenance is a consequence of this policy, not an empirical estimate that every claim lacks source support.
\textbf{Recovered-source checking} ($U_r$) first resolves units to declared spans or, when absent, spans recovered by the same protocol used for grounded citation scoring.
It deduplicates the pooled source IDs before verifying each claim; absent evidence still counts as unsupported.
We evaluate this third mode for RECOMP and LLMLingua-2 at nominal 0.25 on both datasets.
For methods whose units all declare provenance, recovered-source and declared-span checking coincide.
We report generator input, logical verifier input (premise plus hypothesis), and their sum under declared-span checking as a partial token-cost proxy (Table~\ref{tab:cost}).

\paragraph{Evidence selection on QASPER.}
On QASPER \citep{dasigi2021qasper} we additionally evaluate the compressor's own attribution: declared cited sids are mapped to their (whitespace-normalized) full paragraph texts and scored by set-F1 against the gold evidence paragraphs, taking the maximum over annotators per the official protocol.
Runs without declared provenance have no declared-evidence F1; recovered-span scores would measure a different attribution pathway.

\section{Experiments}
\label{sec:experiments}

\subsection{Experimental Setup}
\label{sec:experimental-setup}
We evaluate on two attribution benchmarks with different retrieval granularities.
\textbf{ASQA} comprises 948 ambiguous factoid questions from the ALCE release \citep{gao2023alce}, each paired with the top-5 GTR-retrieved passages, sentence-segmented into citable snippets; answer quality is string exact match (str-EM).
\textbf{QASPER} \citep{dasigi2021qasper} is the 1,005-question dev split over 281 NLP papers, using each supplied paper's abstract and full text.
Gold evidence is paragraph-level, so identity presents paragraphs as citable units while compressors operate on sentence snippets; declared sids map to paragraphs for set-F1 against gold evidence, max over annotators.
The generator is Qwen2.5-7B-Instruct served via vLLM with greedy decoding and ALCE-style numbered-context citation prompting \citep{gao2023alce}.
Primary entailment judgments use the development-grade DeBERTa NLI evaluator of \S\ref{sec:protocol-setup} at $\tau = 0.5$.
We compare six compressors: \emph{identity} (no compression; document-level on ASQA, paragraph-level on QASPER), cross-encoder \emph{rerank top-$k$} (ms-marco-MiniLM-L-6-v2), embedding-based \emph{extractive} sentence selection (all-MiniLM-L6-v2), the \emph{RECOMP} abstractive summarizer (NQ checkpoint) \citep{xu2024recomp}, \emph{LLMLingua-2} token pruning \citep{pan2024llmlingua2}, and our \emph{extract--cluster--rewrite} (ECR) variant, which ranks sentences, greedily clusters near-duplicates by Jaccard word overlap (threshold 0.5, with a negation guard), and emits one unit per cluster---the highest-relevance member sentence verbatim---citing \emph{all} member sentence ids.
Although LLMLingua-2 prunes tokens verbatim, its output is a token stream with no unit-level sid mapping in our pipeline, so it declares no provenance and is scored with the no-provenance family alongside RECOMP.
Each compressor runs at nominal budgets $\{0.5, 0.25, 0.1, 0.05\}$ of the retrieved context, enforced in generator-tokenizer tokens.
Because compressors do not reliably hit their nominal budgets, the frontier figures and primary summary tables report the \emph{achieved} ratio: RECOMP's summary length is nearly budget-independent, saturating at achieved ratios of 0.05--0.08 on ASQA and $\approx$0.015 on QASPER across all four nominal budgets.
All primary comparisons use the same successful questions across all 21 settings: 944 on ASQA and 976 from 271 papers on QASPER.
Only identity has failures: 4 budget-check failures on ASQA, and 15 budget-check plus 14 prompt-overflow failures on QASPER; their question IDs are excluded from every primary setting.
Full-cohort descriptive aggregates remain available in the evaluation artifacts.

\paragraph{QASPER context and token counts.}
The context contains the supplied paper's abstract and all full-text paragraphs, sentence-segmented for compression; no top-$k$ retrieval truncation is applied.
On the 976 common questions, full-source token counts have mean 4,672, median 4,557, and range 897--11,418.
At nominal 0.25, RECOMP's mean compressed context is 60.3 tokens and its mean complete generator input is 125.4 tokens.
The reported achieved fraction, 0.01525, is the mean of per-question context/full-source ratios, not the ratio of these token means.
At nominal 0.50, this gap between nominal and achieved fractions is about 32.8-fold.

\subsection{Statistical Analysis and Metric Audit}
\label{sec:statistics}

\paragraph{Common cohorts and estimands.}
Primary summaries and comparisons use the intersection of successful question IDs across identity and all twenty compressed settings.
Each question has equal weight in citation precision, citation recall, answer quality, and evidence F1; claims and citations are not pooled across questions.
Undefined evidence or recovery metrics remain undefined rather than becoming zeros.
Verification-rate means exclude answers with zero claims; paired verification comparisons exclude a question if either member has zero claims.
The analysis artifacts report those exclusions and retain full-cohort descriptive means separately.

\paragraph{Inference.}
We use pointwise 95\% percentile bootstrap intervals for each paired mean difference \citep{dror2018testing}, resampling 944 questions on ASQA and 271 papers on QASPER with replacement.
All retained questions from a sampled QASPER paper travel together, and each replicate divides its total difference by the number of sampled questions, preserving the question-weighted estimand.
Every QASPER contrast retains all 271 papers after metric-specific exclusions.
Two-sided randomization tests swap paired labels within questions on ASQA and jointly within papers on QASPER.
These tests assume label exchangeability within a resampling unit under the null.
Both procedures use 10,000 replicates and seed 13, with independent random-number streams for bootstrap sampling and randomization.
Monte Carlo $p$-values use $(b+1)/(10{,}000+1)$, where $b$ counts randomizations at least as extreme as the observed absolute difference.
The initial plan specified 134 core contrasts and 10 recovery-threshold contrasts before inspecting the revised inferential results.
After that analysis and a cache-coverage audit, we added four recovered-source versus summary-verification contrasts at nominal 0.25 for RECOMP and LLMLingua-2 on both datasets.
The extension plan was recorded before running these four tests, and Holm adjustment was recomputed over all 148 raw $p$-values.
We denote Holm-adjusted values by $p_H$.
All reported adjusted values use this expanded family; original effects, intervals, and raw $p$-values are unchanged.
This is a post hoc camera-ready analysis, not a preregistration.
Intervals are pointwise, not simultaneous; statistical-support statements use $p_H<0.05$.
Failure to reject a difference does not establish equivalence or preservation.
All inference is conditional on the fixed generator outputs, recovery procedure, and evaluator.

\paragraph{Precision-code audit.}
The manuscript's precision definition requires the concatenated cited evidence to entail a claim before any individual citation receives credit.
A code audit found that the previous implementation could credit an individually supporting citation even when joint entailment failed.
A saved ASQA answer reproduced this discrepancy, and a regression test now checks the required gate through the evaluation pipeline.
All saved answers are rescored with the corrected definition using the same DeBERTa checkpoint in float32; no answers are regenerated.
The audit also computes legacy precision from the same new NLI scores to isolate the gate's effect from numerical inference differences and cohort selection.
On the common ASQA cohort, the gate reduces identity precision from 0.5622 to 0.5086, affecting 98 of 944 questions.
On QASPER it reduces identity precision from 0.4992 to 0.4826, affecting 61 of 976 questions.
These are deterministic scorer-rule comparisons using fixed NLI scores, not new model-performance tests.
Input, source-code, checkpoint, and tokenizer hashes accompany persistent per-question records and resumable score caches.
These primary scores remain dependent on this single evaluator; the correction does not constitute independent validation.
The 512-token NLI input limit can truncate concatenated evidence, so the joint gate is also conditional on that bounded input.

Appendix~\ref{app:paired-comparisons} reports all 148 tested differences, including those not retained as claims; structural provenance rejections are descriptive.

\begin{figure*}[t]
  \centering
  \includegraphics[width=\textwidth]{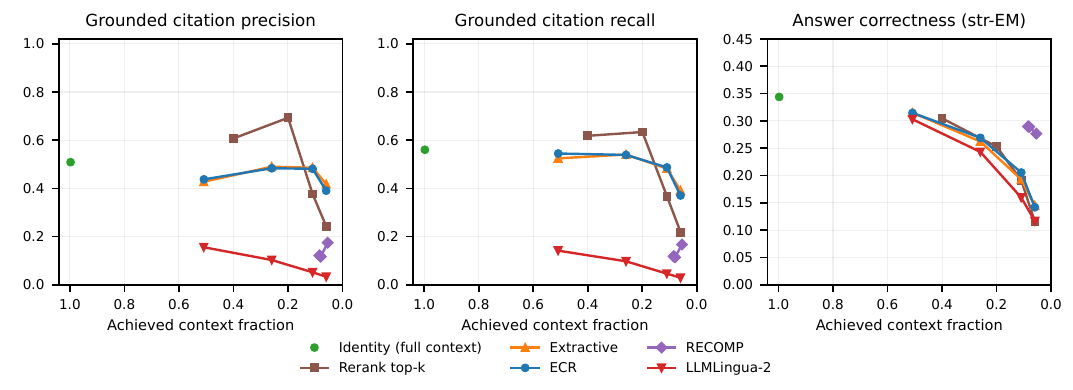}
  \caption{The attribution--compression frontier on 944 common ASQA questions. Compression increases rightward ($x$: achieved fraction of retrieved context). Curves show descriptive means under the shared evaluator and recovery protocol; paired difference intervals and tests appear in Appendix~\ref{app:paired-comparisons}. The right panel reports str-EM.}
  \label{fig:frontier-asqa}
\end{figure*}

\begin{figure*}[t]
  \centering
  \includegraphics[width=\textwidth]{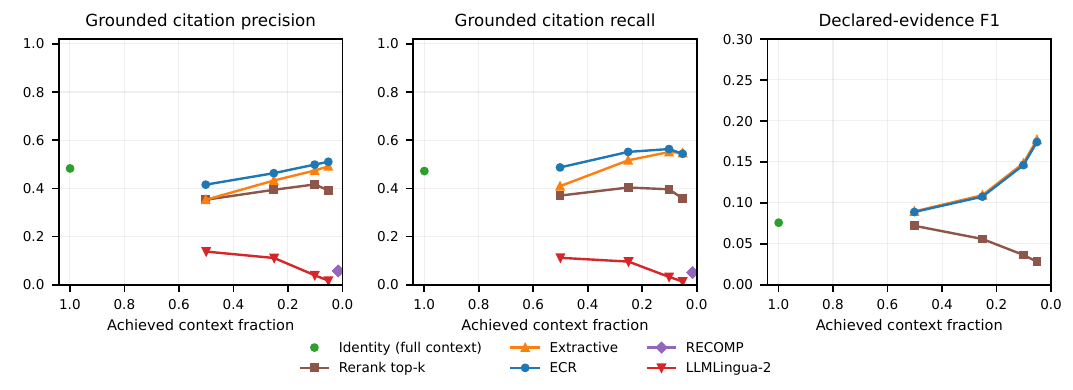}
  \caption{The frontier on 976 common QASPER questions from 271 papers, axes as in Figure~\ref{fig:frontier-asqa}. The third panel scores each compressor's \emph{declared} evidence against gold paragraphs; it is undefined for no-provenance methods. Small plotted differences do not establish equivalence.}
  \label{fig:frontier-qasper}
\end{figure*}

% AUTO-GENERATED by scripts/build_step2_tables.py. DO NOT EDIT.
\begin{table}[t]
\centering
\small
\setlength{\tabcolsep}{3pt}
\begin{tabular}{lcccccc}
\toprule
Compressor & $\hat{\rho}$ & $P_g$ & $P_e$ & $U_s$ & $U_m$ & $U_r$ \\
\midrule
\multicolumn{7}{l}{\emph{ASQA}} \\
Identity & 1.00 & 0.51 & 0.51 & 0.44 & 0.44 & 0.44 \\
Rerank top-$k$ & 0.20 & 0.69 & 0.69 & 0.36 & 0.37 & 0.36 \\
Extractive & 0.26 & 0.49 & 0.49 & 0.46 & 0.46 & 0.46 \\
ECR & 0.26 & 0.48 & 0.49 & 0.46 & 0.45 & 0.46 \\
RECOMP & 0.08 & 0.12 & 0.86 & 1.00 & 0.17 & 0.88 \\
LLMLingua-2 & 0.26 & 0.10 & 0.53 & 1.00 & 0.51 & 0.90 \\
\midrule
\multicolumn{7}{l}{\emph{QASPER}} \\
Identity & 1.00 & 0.48 & 0.48 & 0.53 & 0.53 & 0.53 \\
Rerank top-$k$ & 0.25 & 0.39 & 0.39 & 0.59 & 0.60 & 0.59 \\
Extractive & 0.25 & 0.43 & 0.43 & 0.48 & 0.48 & 0.48 \\
ECR & 0.25 & 0.46 & 0.46 & 0.45 & 0.45 & 0.45 \\
RECOMP & 0.02 & 0.06 & 0.55 & 1.00 & 0.50 & 0.95 \\
LLMLingua-2 & 0.25 & 0.11 & 0.42 & 1.00 & 0.65 & 0.91 \\
\bottomrule
\end{tabular}
\caption{Citation precision and unsupported rates at nominal budget 0.25 (identity: full context), on 944 ASQA and 976 QASPER questions. Notation follows Section~\ref{sec:protocol}; no-provenance declared-span rates are structural. Two zero-claim RECOMP answers are excluded from QASPER verification rates.}
\label{tab:laundering}
\end{table}

\section{Results}
\label{sec:results}

\paragraph{Emitted and grounded scores diverge (H1).}
Figures~\ref{fig:frontier-asqa} and~\ref{fig:frontier-qasper} show the frontier.
On ASQA at nominal budget 0.25 (achieved ratio 0.08), RECOMP receives \emph{emitted} citation precision of 0.86 against its summaries and \emph{grounded} precision of 0.12 against source spans; its mean per-question recovery rate is 0.41.
The paired precision gap is 0.740 (95\% CI [0.711, 0.767], $p_H=0.0148$).
The gap combines effects of compression, bounded recovery, and NLI judgments; these measurements do not separate their contributions.
On QASPER the corresponding scores are 0.55 and 0.06 (achieved ratio 0.015), with gap 0.491 [0.457, 0.525] and $p_H=0.0148$.
LLMLingua-2's grounded precision means decrease across budgets: 0.16, 0.10, 0.05, 0.03 on ASQA and 0.14, 0.11, 0.04, 0.02 on QASPER.
All no-provenance settings score below their dataset's identity reference (0.51 on ASQA, 0.48 on QASPER), with $p_H=0.0148$ for each comparison.

\paragraph{Extraction and reranking trade off citation scores and answer quality.}
ASQA extraction yields grounded precision 0.43, 0.49, 0.49, and 0.42 across achieved ratios 0.51, 0.26, 0.11, and 0.06.
Relative to identity, precision is lower at nominal 0.50 and 0.05 ($p_H=0.0148$); the intermediate differences are unresolved ($p_H=1$), not evidence of preservation.
Its str-EM decreases from identity's 0.34 through 0.32, 0.26, and 0.19 to 0.15, with every adjacent decrease supported ($p_H=0.0148$).
ASQA reranking at nominal 0.25 increases precision from 0.51 to 0.69 ($\Delta=0.184$, CI [0.149, 0.219], $p_H=0.0148$), while recall rises from 0.56 to 0.63 and str-EM falls from 0.34 to 0.25 (both $p_H=0.0148$).
The corresponding QASPER precision difference is negative ($\Delta=-0.089$, CI [$-0.124$, $-0.052$], $p_H=0.0148$).
These dataset-specific results do not isolate a causal effect of citation granularity.

\paragraph{Declared provenance determines verification coverage (H4).}
Table~\ref{tab:laundering} contrasts declared-span and summary checking.
For RECOMP on ASQA the unsupported rates are 1.00 and 0.17, respectively; the corresponding QASPER values are 1.00 and 0.50.
LLMLingua-2 has rates of 1.00 vs.\ 0.51 on ASQA and 1.00 vs.\ 0.65 on QASPER.
Every $1.00$ here is structural: no declared source evidence reaches the span verifier, so claims fail without an entailment judgment.
These values do not establish that all claims lack recoverable source support.
Recovered-source verification now supplies a direct claim-level comparison: for RECOMP, unsupported rates are 0.88 on ASQA and 0.95 on QASPER.
Their paired differences from summary checking are 0.712 (CI [0.683, 0.739]) and 0.450 [0.417, 0.483], both $p_H=0.0148$.
LLMLingua-2's recovered-source rates are 0.90 and 0.91, with differences 0.394 [0.362, 0.426] and 0.253 [0.225, 0.282] (both $p_H=0.0148$).
These gaps persist when recovery is attempted, but remain conditional on the bounded search and shared NLI evaluator.
For provenance-declaring methods, displayed rate differences are below 0.01 in magnitude; no mode comparison survives Holm correction, which does not establish equivalence.

\paragraph{ECR's advantage is limited to one tested setting.}
On QASPER, ECR improves grounded precision over extraction at nominal 0.50 ($\Delta=0.063$, CI [0.033, 0.092], $p_H=0.0148$).
Its other QASPER precision differences and all four ASQA differences lack adjusted statistical support.
The tightest QASPER ECR setting also does not establish an advantage over identity ($\Delta=0.028$, CI [$-0.010$, 0.064], $p_H=1$).
ECR's QASPER declared-evidence F1 is slightly lower than extraction at every budget, by 0.00065--0.00295 (all $p_H<0.05$).
Thus the results support a setting-specific precision gain, not general superiority or lossless redundancy removal.

\section{Analysis}
\label{sec:analysis}

\paragraph{Re-attribution varies across no-provenance methods.}
The re-attribution rate measures whether at least one candidate source snippet passes the NLI entailment threshold for a unit without declared provenance.
On ASQA, RECOMP's rates are 0.41, 0.41, 0.42, and 0.39 across nominal budgets 0.5 through 0.05, excluding examples without summary units.
Averaged over these examples, roughly three-fifths of its units have no passing candidate; this does not establish that the source contains no supporting evidence.
LLMLingua-2's ASQA rate means decrease from 0.56 to 0.40, 0.25, and 0.20 as budgets tighten; the first two adjacent decreases have $p_H=0.0148$, while the final decrease is unresolved ($p_H=0.242$).
Recovery coverage depends on method, budget, and dataset as well as the search and evaluator; these observations provide no universal ceiling on recoverable support.

\paragraph{Verification adds input tokens.}
% AUTO-GENERATED by scripts/build_step2_tables.py. DO NOT EDIT.
\begin{table}[t]
\centering
\small
\setlength{\tabcolsep}{3pt}
\begin{tabular}{lrrrr}
\toprule
Compressor & Gen.\ in & Ver.\ span & Ver.\ sum. & Total \\
\midrule
Identity & 798 & 371 & 382 & 1169 \\
Rerank top-$k$ & 210 & 171 & 174 & 381 \\
Extractive & 277 & 112 & 114 & 389 \\
ECR & 277 & 118 & 107 & 395 \\
RECOMP & 127 & 0 & 84 & 127 \\
LLMLingua-2 & 253 & 0 & 240 & 253 \\
\bottomrule
\end{tabular}
\caption{Mean input tokens on 944 ASQA questions at nominal budget 0.25 (identity: full context). Total includes generator and declared-span verification input; compression, output, and recovery are excluded.}
\label{tab:cost}
\end{table}

Table~\ref{tab:cost} reports a generator-input-plus-verifier-input proxy, measured in generator-tokenizer tokens before NLI truncation.
It excludes compressor inference, generated output, recovery, and differences in model-specific tokenization or per-token cost.
The declared-span column is zero for no-provenance compressors because this policy rejects their claims without an NLI call.
That zero does not describe the cost of recovering and checking source evidence.

\paragraph{Evidence selection under pressure.}
The declared-evidence panel of Figure~\ref{fig:frontier-qasper} shows increasing F1 for extraction and ECR as budgets tighten, from about 0.09 to 0.18 and 0.17, respectively.
Each adjacent increase is supported ($p_H=0.0148$).
Reranking's F1 means decrease from 0.07 to 0.06, 0.04, and 0.03; the first two decreases have $p_H=0.0148$, but the final difference is unresolved ($p_H=1$).
These are evidence-F1 results, not evidence-precision estimates or a controlled test of selection granularity.

\paragraph{Sensitivity to the re-attribution threshold.}
\label{par:tau}
The re-attributability protocol depends on the entailment threshold $\tau$.
Table~\ref{tab:tau} sweeps $\tau$ from 0.3 to 0.8 for both no-provenance
methods on ASQA (citation scoring fixed at the production threshold of
0.5).
Grounded precision ranges from 0.15 to 0.08 for RECOMP and from 0.16 to 0.04 for LLMLingua-2, while recovery rates fall from 0.48 to 0.31 and from 0.54 to 0.20, respectively.
Emitted precision exceeds grounded precision at every tested recovery threshold for both methods ($p_H=0.0148$ throughout).
This is sensitivity analysis of one operating choice, not independent validation: the evaluator, top-10 lexical search, and single-snippet support requirement remain fixed.

% AUTO-GENERATED by scripts/build_step2_tables.py. DO NOT EDIT.
\begin{table}[t]
\centering
\small
\setlength{\tabcolsep}{3pt}
\begin{tabular}{ccccc}
\toprule
& \multicolumn{2}{c}{RECOMP} & \multicolumn{2}{c}{LLMLingua-2} \\
\cmidrule(lr){2-3}\cmidrule(lr){4-5}
$\tau$ & $P_g$ & reattr. & $P_g$ & reattr. \\
\midrule
0.3 & 0.15 & $0.48^{n=929}$ & 0.16 & 0.54 \\
0.4 & 0.13 & $0.44^{n=929}$ & 0.13 & 0.47 \\
0.5 & 0.12 & $0.41^{n=929}$ & 0.10 & 0.40 \\
0.6 & 0.11 & $0.38^{n=929}$ & 0.08 & 0.34 \\
0.7 & 0.09 & $0.34^{n=929}$ & 0.05 & 0.27 \\
0.8 & 0.08 & $0.31^{n=929}$ & 0.04 & 0.20 \\
\bottomrule
\end{tabular}
\caption{Recovery-threshold sensitivity on 944 ASQA questions at nominal budget 0.25; citation scoring stays at $\tau=0.5$. $P_g$: grounded precision; reattr.: re-attribution rate. Superscripts count defined values; dashes mark undefined metrics.}
\label{tab:tau}
\end{table}

\paragraph{Second-evaluator sensitivity.}
We rescore 100 questions per dataset with TRUE T5-XXL \citep{honovich2022true}, covering full-context identity plus extraction, RECOMP, and LLMLingua-2 at nominal 0.25.
With source mappings fixed, RECOMP's emitted/grounded precision is 0.78/0.13 on ASQA and 0.44/0.01 on QASPER.
The paired gaps are 0.65 [0.55, 0.75] and 0.43 [0.33, 0.53], respectively.
When TRUE also selects recovered spans, grounded precision is 0.14 and 0.03.
All 16 planned emitted-minus-grounded precision and recall gaps across both methods, datasets, and recovery modes are positive ($p_H=0.0024$ in a separate 24-test family).
None of the eight cross-evaluator grounded-precision differences passes Holm adjustment; this does not establish equivalence or validate the primary numbers.
Appendix~\ref{app:true-audit} reports all comparisons, the 4,096-token input configuration, decoder anomalies, and documented restarts.
The pattern persists under a second scoring family, with recovery and calibration limitations retained.

\section{Discussion}
\label{sec:discussion}

\paragraph{What laundering looks like.}
Consider an ASQA example under RECOMP at nominal budget 0.25.
Asked \emph{``How many state parks are there in Virginia?''}, the generator answers \emph{``There are 34 state parks in Virginia [1]''}, while Document [1] states \emph{``Virginia has 34 state parks and 17 state forests\,\ldots''}.
The citation receives full emitted precision, but Document [1] is the compressor's summary with no declared source link.
In another example, the answer \emph{``The man who killed [them] is not mentioned in the retrieved documents [1]''} cites a summary containing that same assertion of absence.
Both illustrate consistency with intermediate text without a declared path to source evidence; neither example alone establishes an independently verified hallucination.

\paragraph{Implications for practice.}
Five operational recommendations follow from this evaluation setup.
(1)~\emph{Verify against source spans:} consistency with compressor output does not establish support from locatable evidence.
Report whether spans are declared or recovered and how support is judged.
(2)~\emph{Report achieved compression:} nominal fractions overstate achieved context size by up to 32.8-fold here (Section~\ref{sec:experimental-setup}).
(3)~\emph{Include verification in cost reporting:} report generator and verifier input lengths together, while identifying which other costs remain unmeasured (Table~\ref{tab:cost}).
Zero declared-span cost can mean that no evidence reached the verifier.
(4)~\emph{Evaluate post-hoc recovery explicitly:} missed recovery can reflect search errors or missing support (Section~\ref{sec:analysis}).
(5)~\emph{Document citable-unit granularity:} comparisons differ in their source units as well as their datasets.
Controlled ablations are needed to isolate the effect of unit size.

\paragraph{What dominating the frontier would require.}
Future compressors should merge redundant evidence while retaining validated source links and accounting for verification cost.
ECR's limited measured gains motivate independent calibration and replication before claiming general improvements.

\paragraph{Failure modes and excluded examples.}
All 33 failures occur in identity runs: 4 budget-check failures on ASQA, and 15 budget-check plus 14 prompt-overflow failures on QASPER.
No compressed run fails on these examples.
Primary comparisons exclude the same failed question IDs from every setting, so their estimand is performance conditional on identity success.

\section{Conclusion}

We measured citation attribution across context-compression methods and budgets under a fixed generator and primary NLI evaluator.
Emitted and grounded scores diverge, but their magnitudes depend on recovery and evaluation choices that have not been independently calibrated.
Declared-span verification rejects claims lacking declared evidence by policy; it does not determine whether support could be recovered.
Paired comparisons quantify sampling uncertainty conditional on the fixed outputs and evaluator, with only limited evidence for ECR's advantage.
We recommend reporting grounded alongside emitted citation scores and achieved alongside nominal compression.
The TRUE subsample retains the emitted--grounded gaps; human calibration and broader replication remain necessary.

\section{Limitations}

\paragraph{Evaluator dependence.}
Primary scores use DeBERTa-v3-base, trained on MNLI, FEVER, and ANLI, at $\tau=0.5$, without human calibration.
Using the same model to select recovered spans and score support can reinforce systematic errors; the resulting scores are evaluator-dependent estimates, not validated rates of source support or factual correctness.
This dependence also limits conclusions about method rankings and the size of emitted--grounded gaps.
The 200-question TRUE audit checks a second scoring family, but fixed recovery retains DeBERTa selection and TRUE recovery reuses one judge.
Different input windows and nonbinary decoder outputs further limit numerical comparison.
Human source-level calibration remains necessary \citep{rashkin2023ais}.

\paragraph{Single generator.}
All results use Qwen2.5-7B-Instruct with greedy decoding. Frontier shapes may differ across model families and scales; a second-family replication is planned.

\paragraph{Dataset coverage.}
We evaluate two English datasets (ASQA and QASPER) with distinct retrieval granularities, but this does not cover multilingual settings, open-web retrieval, or longer-context regimes.

\paragraph{Re-attribution hyperparameters.}
The top-10 lexical candidate pool can miss semantically matching snippets, and requiring one snippet to entail a unit can miss support distributed across snippets.
Consequently, a failed recovery does not establish that the source contains no support.
Table~\ref{tab:tau} varies the recovery threshold while holding the scoring threshold fixed; it does not validate the evaluator or test candidate-pool size, retrieval strategy, or multi-snippet recovery.

\paragraph{Verification scope.}
Declared-span verification rejects claims lacking declared evidence by construction, even if support could be recovered.
Its $1.00$ unsupported rate for no-provenance runs is a provenance-availability diagnostic, not a calibrated hallucination rate.
Recovered-source verification is reported at nominal 0.25 for the two no-provenance methods and remains dependent on the same bounded recovery procedure and evaluator.

\paragraph{Residual error rows.}
Only identity fails on 4 ASQA and 29 QASPER questions; every primary setting is evaluated on the resulting common successful subset.
This pairing prevents comparisons from using different questions, but conditions results on identity success and omits those difficult cases.
Sampling intervals condition on the fixed generated outputs and evaluator; they do not include systematic NLI error or variation across generation runs.

\paragraph{Nominal versus achieved budgets.}
Abstractive compressors do not reliably hit nominal token budgets; RECOMP often produces much less context than requested.
We report \emph{achieved} compression ratios, but comparisons at a shared nominal budget do not establish performance at matched achieved context sizes.

\section*{Reproducibility Statement}
Saved generation outputs, input hashes, and pinned evaluator revisions accompany per-question rescoring records.
The fixed contrast plans and paired-analysis reports regenerate the numerical tables and frontier figures without further model calls.
The harness, configurations, protocol, and per-question records will be publicly released upon publication.

\clearpage
\bibliography{references}

\appendix
\twocolumn[\section*{Appendices}]
\section{Results on Common Questions}
\label{app:full-results}

Tables~\ref{tab:full-asqa} and~\ref{tab:full-qasper} report all primary settings.
$\rho$: nominal budget; $\hat{\rho}$: achieved ratio;
$P_g/R_g$: grounded citation precision/recall; $P_e$:
emitted citation precision; reattr.: re-attribution rate (units without declared provenance recovering at
least one source span under the NLI protocol); $U_s/U_m$: unsupported-claim
rate under declared-span-only and summary-checked verification; str-EM: answer string exact match (ASQA); EF1: declared-evidence F1
(QASPER only).
Every setting uses the same 944 ASQA or 976 QASPER successful questions; $n$ denotes that common cohort, with metric-specific omissions identified in the tables.
Scores are question-weighted means over defined values; superscript $n=\cdots$ gives the nonmissing count when smaller than the cohort, and dashes mark undefined metrics.
QASPER verification rates exclude 1, 2, 3, and 2 zero-claim RECOMP answers at nominal budgets 0.50, 0.25, 0.10, and 0.05, respectively; other settings use the full cohort.
The $1.00$ declared-span rates for no-provenance methods are structural; this verification mode does not attempt recovery.
Exploratory slot-compressor runs are excluded.
RECOMP emits identical per-question contexts at nominal budgets 0.50 and 0.25 on ASQA, and at 0.50, 0.25, and 0.10 on QASPER.
Separate generation passes are retained as distinct operating points.

% AUTO-GENERATED by scripts/build_step2_tables.py. DO NOT EDIT.
\begin{table*}[t]
\centering
\small
\setlength{\tabcolsep}{3pt}
\begin{tabular}{lcccccccccc}
\toprule
Compressor & $\rho$ & $\hat{\rho}$ & $P_g$ & $R_g$ & $P_e$ & reattr. & str-EM & $U_s$ & $U_m$ & $n$ \\
\midrule
Identity & 1 & 1.00 & 0.51 & 0.56 & 0.51 & -- & 0.34 & 0.44 & 0.44 & 944 \\
\addlinespace[2pt]
Rerank top-$k$ & 0.5 & 0.40 & 0.61 & 0.62 & 0.61 & -- & 0.30 & 0.38 & 0.38 & 944 \\
Rerank top-$k$ & 0.25 & 0.20 & 0.69 & 0.63 & 0.69 & -- & 0.25 & 0.36 & 0.37 & 944 \\
Rerank top-$k$ & 0.1 & 0.11 & 0.38 & 0.37 & 0.38 & -- & 0.19 & 0.63 & 0.63 & 944 \\
Rerank top-$k$ & 0.05 & 0.06 & 0.24 & 0.22 & 0.24 & -- & 0.12 & 0.78 & 0.78 & 944 \\
\addlinespace[2pt]
Extractive & 0.5 & 0.51 & 0.43 & 0.52 & 0.43 & -- & 0.32 & 0.47 & 0.48 & 944 \\
Extractive & 0.25 & 0.26 & 0.49 & 0.54 & 0.49 & -- & 0.26 & 0.46 & 0.46 & 944 \\
Extractive & 0.1 & 0.11 & 0.49 & 0.48 & 0.49 & -- & 0.19 & 0.52 & 0.52 & 944 \\
Extractive & 0.05 & 0.06 & 0.42 & 0.39 & 0.42 & -- & 0.15 & 0.61 & 0.61 & 944 \\
\addlinespace[2pt]
ECR & 0.5 & 0.51 & 0.44 & 0.54 & 0.45 & -- & 0.31 & 0.45 & 0.44 & 944 \\
ECR & 0.25 & 0.26 & 0.48 & 0.54 & 0.49 & -- & 0.27 & 0.46 & 0.45 & 944 \\
ECR & 0.1 & 0.11 & 0.48 & 0.49 & 0.49 & -- & 0.21 & 0.51 & 0.51 & 944 \\
ECR & 0.05 & 0.06 & 0.39 & 0.37 & 0.39 & -- & 0.14 & 0.63 & 0.63 & 944 \\
\addlinespace[2pt]
RECOMP & 0.5 & 0.08 & 0.12 & 0.12 & 0.86 & $0.41^{n=929}$ & 0.29 & 1.00 & 0.17 & 944 \\
RECOMP & 0.25 & 0.08 & 0.12 & 0.12 & 0.86 & $0.41^{n=929}$ & 0.29 & 1.00 & 0.17 & 944 \\
RECOMP & 0.1 & 0.08 & 0.12 & 0.11 & 0.86 & $0.42^{n=929}$ & 0.29 & 1.00 & 0.17 & 944 \\
RECOMP & 0.05 & 0.05 & 0.17 & 0.17 & 0.80 & $0.39^{n=928}$ & 0.28 & 1.00 & 0.24 & 944 \\
\addlinespace[2pt]
LLMLingua-2 & 0.5 & 0.51 & 0.16 & 0.14 & 0.58 & 0.56 & 0.30 & 1.00 & 0.45 & 944 \\
LLMLingua-2 & 0.25 & 0.26 & 0.10 & 0.10 & 0.53 & 0.40 & 0.24 & 1.00 & 0.51 & 944 \\
LLMLingua-2 & 0.1 & 0.11 & 0.05 & 0.05 & 0.31 & 0.25 & 0.16 & 1.00 & 0.73 & 944 \\
LLMLingua-2 & 0.05 & 0.06 & 0.03 & 0.03 & 0.16 & 0.20 & 0.12 & 1.00 & 0.86 & 944 \\
\bottomrule
\end{tabular}
\caption{All 21 primary settings on 944 common ASQA questions; question-weighted means. Notation and exclusions follow Appendix~\ref{app:full-results}. No-provenance declared-span rates are structural.}
\label{tab:full-asqa}
\end{table*}

% AUTO-GENERATED by scripts/build_step2_tables.py. DO NOT EDIT.
\begin{table*}[t]
\centering
\small
\setlength{\tabcolsep}{3pt}
\begin{tabular}{lcccccccccc}
\toprule
Compressor & $\rho$ & $\hat{\rho}$ & $P_g$ & $R_g$ & $P_e$ & reattr. & EF1 & $U_s$ & $U_m$ & $n$ \\
\midrule
Identity & 1 & 1.00 & 0.48 & 0.47 & 0.48 & -- & 0.08 & 0.53 & 0.53 & 976 \\
\addlinespace[2pt]
Rerank top-$k$ & 0.5 & 0.50 & 0.35 & 0.37 & 0.35 & -- & 0.07 & 0.63 & 0.63 & 976 \\
Rerank top-$k$ & 0.25 & 0.25 & 0.39 & 0.40 & 0.39 & -- & 0.06 & 0.59 & 0.60 & 976 \\
Rerank top-$k$ & 0.1 & 0.10 & 0.42 & 0.40 & 0.42 & -- & 0.04 & 0.60 & 0.60 & 976 \\
Rerank top-$k$ & 0.05 & 0.05 & 0.39 & 0.36 & 0.39 & -- & 0.03 & 0.64 & 0.64 & 976 \\
\addlinespace[2pt]
Extractive & 0.5 & 0.50 & 0.35 & 0.41 & 0.35 & -- & 0.09 & 0.59 & 0.59 & 976 \\
Extractive & 0.25 & 0.25 & 0.43 & 0.52 & 0.43 & -- & 0.11 & 0.48 & 0.48 & 976 \\
Extractive & 0.1 & 0.10 & 0.47 & 0.55 & 0.47 & -- & 0.15 & 0.44 & 0.45 & 976 \\
Extractive & 0.05 & 0.05 & 0.49 & 0.55 & 0.49 & -- & 0.18 & 0.45 & 0.45 & 976 \\
\addlinespace[2pt]
ECR & 0.5 & 0.50 & 0.42 & 0.49 & 0.42 & -- & 0.09 & 0.51 & 0.51 & 976 \\
ECR & 0.25 & 0.25 & 0.46 & 0.55 & 0.46 & -- & 0.11 & 0.45 & 0.45 & 976 \\
ECR & 0.1 & 0.10 & 0.50 & 0.56 & 0.50 & -- & 0.15 & 0.43 & 0.44 & 976 \\
ECR & 0.05 & 0.05 & 0.51 & 0.54 & 0.51 & -- & 0.17 & 0.45 & 0.46 & 976 \\
\addlinespace[2pt]
RECOMP & 0.5 & 0.02 & 0.06 & 0.05 & 0.56 & $0.44^{n=845}$ & -- & $1.00^{n=975}$ & $0.49^{n=975}$ & 976 \\
RECOMP & 0.25 & 0.02 & 0.06 & 0.05 & 0.55 & $0.44^{n=845}$ & -- & $1.00^{n=974}$ & $0.50^{n=974}$ & 976 \\
RECOMP & 0.1 & 0.02 & 0.06 & 0.05 & 0.55 & $0.44^{n=845}$ & -- & $1.00^{n=973}$ & $0.50^{n=973}$ & 976 \\
RECOMP & 0.05 & 0.02 & 0.06 & 0.05 & 0.55 & $0.44^{n=845}$ & -- & $1.00^{n=974}$ & $0.50^{n=974}$ & 976 \\
\addlinespace[2pt]
LLMLingua-2 & 0.5 & 0.50 & 0.14 & 0.11 & 0.50 & 0.92 & -- & 1.00 & 0.61 & 976 \\
LLMLingua-2 & 0.25 & 0.25 & 0.11 & 0.10 & 0.42 & 0.79 & -- & 1.00 & 0.65 & 976 \\
LLMLingua-2 & 0.1 & 0.10 & 0.04 & 0.03 & 0.43 & 0.35 & -- & 1.00 & 0.63 & 976 \\
LLMLingua-2 & 0.05 & 0.05 & 0.02 & 0.01 & 0.38 & 0.18 & -- & 1.00 & 0.67 & 976 \\
\bottomrule
\end{tabular}
\caption{All 21 primary settings on 976 common QASPER questions; question-weighted means. Notation and exclusions follow Appendix~\ref{app:full-results}. No-provenance declared-span rates are structural.}
\label{tab:full-qasper}
\end{table*}

\newpage
\section{Complete Paired Comparisons}
\label{app:statistics}
\label{app:paired-comparisons}

Section~\ref{sec:statistics} describes the common cohorts, estimands, inference, and precision-code correction.
Tables~\ref{tab:paired-01}--\ref{tab:paired-05} report all 148 planned contrasts, including those not retained as claims; EF1 effects and intervals use four decimal places, and other effects and intervals use three.

\paragraph{Table notation.}
A/Q denote ASQA/QASPER; Id, RR, X, RC, and LL denote identity, reranking, extraction, RECOMP, and LLMLingua-2.
Setting labels give compressor/nominal budget, optionally followed by the recovery threshold.
$P/R$ denote precision/recall, with $e/g$ for emitted/grounded; EM is str-EM, EF1 is evidence F1, and Rec is re-attribution rate.
$U_s/U_m/U_r$ denote declared-span, summary, and recovered-source unsupported rates.
$\Delta$ is the paired left-minus-right mean difference; a single endpoint applies to both settings, and a pair lists left-minus-right metrics.
CI is a pointwise 95\% bootstrap interval; $p_H$ is the two-sided randomization $p$-value after Holm adjustment across all 148 comparisons; $n$ counts included questions.

\paragraph{Recovered-source verification.}
This extension reads existing NLI scores from the sealed cache and makes no new model calls.
It uses the same recovered source IDs as citation scoring, but the claim verifier removes repeated source IDs before pooling evidence.
Consequently, its unsupported rate need not equal one minus grounded citation recall, which retains repeated cited evidence.
Both summary and recovered-source rates exclude the same zero-claim answers: two QASPER RECOMP answers at nominal 0.25; the other three comparisons use their full common cohorts.
Input hashes, the extension plan, cache lookups, per-question records, and all four comparisons are retained separately from the original analysis.

% AUTO-GENERATED by scripts/build_step2_tables.py. DO NOT EDIT.
% Paired-analysis methods and complete-family reporting are documented in sections/statistics.tex.

% AUTO-GENERATED by scripts/build_step2_tables.py. DO NOT EDIT.
\begin{table*}[t]
\centering
\small
\setlength{\tabcolsep}{3pt}
\begin{tabular}{ll ll c r c r r}
\toprule
ID & Data & Left & Right & Endpoint & $\Delta$ & 95\% CI & $p_H$ & $n$ \\
\midrule
% C001: gap_asqa_recomp_50
C001 & A & RC/.5 & RC/.5 & $P_e-P_g$ & $+0.744$ & $[+0.715,+0.771]$ & 0.0148 & 944 \\
% C002: gap_asqa_recomp_25
C002 & A & RC/.25 & RC/.25 & $P_e-P_g$ & $+0.740$ & $[+0.711,+0.767]$ & 0.0148 & 944 \\
% C003: gap_asqa_recomp_10
C003 & A & RC/.1 & RC/.1 & $P_e-P_g$ & $+0.743$ & $[+0.715,+0.770]$ & 0.0148 & 944 \\
% C004: gap_asqa_recomp_5
C004 & A & RC/.05 & RC/.05 & $P_e-P_g$ & $+0.621$ & $[+0.588,+0.654]$ & 0.0148 & 944 \\
% C005: gap_asqa_lingua_50
C005 & A & LL/.5 & LL/.5 & $P_e-P_g$ & $+0.426$ & $[+0.394,+0.458]$ & 0.0148 & 944 \\
% C006: gap_asqa_lingua_25
C006 & A & LL/.25 & LL/.25 & $P_e-P_g$ & $+0.425$ & $[+0.391,+0.458]$ & 0.0148 & 944 \\
% C007: gap_asqa_lingua_10
C007 & A & LL/.1 & LL/.1 & $P_e-P_g$ & $+0.260$ & $[+0.229,+0.291]$ & 0.0148 & 944 \\
% C008: gap_asqa_lingua_5
C008 & A & LL/.05 & LL/.05 & $P_e-P_g$ & $+0.132$ & $[+0.108,+0.156]$ & 0.0148 & 944 \\
% C009: gap_qasper_recomp_50
C009 & Q & RC/.5 & RC/.5 & $P_e-P_g$ & $+0.499$ & $[+0.464,+0.534]$ & 0.0148 & 976 \\
% C010: gap_qasper_recomp_25
C010 & Q & RC/.25 & RC/.25 & $P_e-P_g$ & $+0.491$ & $[+0.457,+0.525]$ & 0.0148 & 976 \\
% C011: gap_qasper_recomp_10
C011 & Q & RC/.1 & RC/.1 & $P_e-P_g$ & $+0.492$ & $[+0.457,+0.526]$ & 0.0148 & 976 \\
% C012: gap_qasper_recomp_5
C012 & Q & RC/.05 & RC/.05 & $P_e-P_g$ & $+0.496$ & $[+0.462,+0.531]$ & 0.0148 & 976 \\
% C013: gap_qasper_lingua_50
C013 & Q & LL/.5 & LL/.5 & $P_e-P_g$ & $+0.360$ & $[+0.319,+0.402]$ & 0.0148 & 976 \\
% C014: gap_qasper_lingua_25
C014 & Q & LL/.25 & LL/.25 & $P_e-P_g$ & $+0.309$ & $[+0.278,+0.341]$ & 0.0148 & 976 \\
% C015: gap_qasper_lingua_10
C015 & Q & LL/.1 & LL/.1 & $P_e-P_g$ & $+0.388$ & $[+0.354,+0.421]$ & 0.0148 & 976 \\
% C016: gap_qasper_lingua_5
C016 & Q & LL/.05 & LL/.05 & $P_e-P_g$ & $+0.361$ & $[+0.329,+0.395]$ & 0.0148 & 976 \\
% C017: gap_asqa_recomp_25_recall
C017 & A & RC/.25 & RC/.25 & $R_e-R_g$ & $+0.712$ & $[+0.684,+0.740]$ & 0.0148 & 944 \\
% C018: identity_asqa_rerank_topk_50
C018 & A & RR/.5 & Id & $P_g$ & $+0.098$ & $[+0.066,+0.131]$ & 0.0148 & 944 \\
% C019: identity_asqa_rerank_topk_25
C019 & A & RR/.25 & Id & $P_g$ & $+0.184$ & $[+0.149,+0.219]$ & 0.0148 & 944 \\
% C020: identity_asqa_rerank_topk_10
C020 & A & RR/.1 & Id & $P_g$ & $-0.132$ & $[-0.170,-0.095]$ & 0.0148 & 944 \\
% C021: identity_asqa_rerank_topk_5
C021 & A & RR/.05 & Id & $P_g$ & $-0.267$ & $[-0.303,-0.230]$ & 0.0148 & 944 \\
% C022: identity_asqa_extractive_50
C022 & A & X/.5 & Id & $P_g$ & $-0.081$ & $[-0.116,-0.047]$ & 0.0148 & 944 \\
% C023: identity_asqa_extractive_25
C023 & A & X/.25 & Id & $P_g$ & $-0.019$ & $[-0.053,+0.016]$ & 1.0000 & 944 \\
% C024: identity_asqa_extractive_10
C024 & A & X/.1 & Id & $P_g$ & $-0.023$ & $[-0.061,+0.014]$ & 1.0000 & 944 \\
% C025: identity_asqa_extractive_5
C025 & A & X/.05 & Id & $P_g$ & $-0.091$ & $[-0.129,-0.053]$ & 0.0148 & 944 \\
% C026: identity_asqa_ecr_50
C026 & A & ECR/.5 & Id & $P_g$ & $-0.071$ & $[-0.105,-0.038]$ & 0.0148 & 944 \\
% C027: identity_asqa_ecr_25
C027 & A & ECR/.25 & Id & $P_g$ & $-0.025$ & $[-0.061,+0.010]$ & 1.0000 & 944 \\
% C028: identity_asqa_ecr_10
C028 & A & ECR/.1 & Id & $P_g$ & $-0.027$ & $[-0.064,+0.010]$ & 1.0000 & 944 \\
% C029: identity_asqa_ecr_5
C029 & A & ECR/.05 & Id & $P_g$ & $-0.118$ & $[-0.155,-0.080]$ & 0.0148 & 944 \\
% C030: identity_asqa_recomp_50
C030 & A & RC/.5 & Id & $P_g$ & $-0.388$ & $[-0.421,-0.354]$ & 0.0148 & 944 \\
\bottomrule
\end{tabular}
\caption{Planned comparisons 1--30 of 148. $\Delta$: left-minus-right mean difference; $p_H$: Holm-adjusted $p$-value. Setting and metric labels follow Appendix~\ref{app:statistics}.}
\label{tab:paired-01}
\end{table*}

% AUTO-GENERATED by scripts/build_step2_tables.py. DO NOT EDIT.
\begin{table*}[t]
\centering
\small
\setlength{\tabcolsep}{3pt}
\begin{tabular}{ll ll c r c r r}
\toprule
ID & Data & Left & Right & Endpoint & $\Delta$ & 95\% CI & $p_H$ & $n$ \\
\midrule
% C031: identity_asqa_recomp_25
C031 & A & RC/.25 & Id & $P_g$ & $-0.387$ & $[-0.421,-0.354]$ & 0.0148 & 944 \\
% C032: identity_asqa_recomp_10
C032 & A & RC/.1 & Id & $P_g$ & $-0.392$ & $[-0.424,-0.359]$ & 0.0148 & 944 \\
% C033: identity_asqa_recomp_5
C033 & A & RC/.05 & Id & $P_g$ & $-0.334$ & $[-0.370,-0.299]$ & 0.0148 & 944 \\
% C034: identity_asqa_lingua_50
C034 & A & LL/.5 & Id & $P_g$ & $-0.353$ & $[-0.386,-0.320]$ & 0.0148 & 944 \\
% C035: identity_asqa_lingua_25
C035 & A & LL/.25 & Id & $P_g$ & $-0.406$ & $[-0.438,-0.374]$ & 0.0148 & 944 \\
% C036: identity_asqa_lingua_10
C036 & A & LL/.1 & Id & $P_g$ & $-0.457$ & $[-0.488,-0.427]$ & 0.0148 & 944 \\
% C037: identity_asqa_lingua_5
C037 & A & LL/.05 & Id & $P_g$ & $-0.475$ & $[-0.504,-0.447]$ & 0.0148 & 944 \\
% C038: identity_qasper_rerank_topk_50
C038 & Q & RR/.5 & Id & $P_g$ & $-0.130$ & $[-0.165,-0.095]$ & 0.0148 & 976 \\
% C039: identity_qasper_rerank_topk_25
C039 & Q & RR/.25 & Id & $P_g$ & $-0.089$ & $[-0.124,-0.052]$ & 0.0148 & 976 \\
% C040: identity_qasper_rerank_topk_10
C040 & Q & RR/.1 & Id & $P_g$ & $-0.066$ & $[-0.104,-0.029]$ & 0.0280 & 976 \\
% C041: identity_qasper_rerank_topk_5
C041 & Q & RR/.05 & Id & $P_g$ & $-0.091$ & $[-0.128,-0.053]$ & 0.0148 & 976 \\
% C042: identity_qasper_extractive_50
C042 & Q & X/.5 & Id & $P_g$ & $-0.130$ & $[-0.159,-0.101]$ & 0.0148 & 976 \\
% C043: identity_qasper_extractive_25
C043 & Q & X/.25 & Id & $P_g$ & $-0.050$ & $[-0.082,-0.019]$ & 0.0832 & 976 \\
% C044: identity_qasper_extractive_10
C044 & Q & X/.1 & Id & $P_g$ & $-0.009$ & $[-0.041,+0.024]$ & 1.0000 & 976 \\
% C045: identity_qasper_extractive_5
C045 & Q & X/.05 & Id & $P_g$ & $+0.009$ & $[-0.027,+0.046]$ & 1.0000 & 976 \\
% C046: identity_qasper_ecr_50
C046 & Q & ECR/.5 & Id & $P_g$ & $-0.067$ & $[-0.099,-0.035]$ & 0.0148 & 976 \\
% C047: identity_qasper_ecr_25
C047 & Q & ECR/.25 & Id & $P_g$ & $-0.020$ & $[-0.050,+0.011]$ & 1.0000 & 976 \\
% C048: identity_qasper_ecr_10
C048 & Q & ECR/.1 & Id & $P_g$ & $+0.016$ & $[-0.019,+0.051]$ & 1.0000 & 976 \\
% C049: identity_qasper_ecr_5
C049 & Q & ECR/.05 & Id & $P_g$ & $+0.028$ & $[-0.010,+0.064]$ & 1.0000 & 976 \\
% C050: identity_qasper_recomp_50
C050 & Q & RC/.5 & Id & $P_g$ & $-0.426$ & $[-0.455,-0.397]$ & 0.0148 & 976 \\
% C051: identity_qasper_recomp_25
C051 & Q & RC/.25 & Id & $P_g$ & $-0.425$ & $[-0.455,-0.396]$ & 0.0148 & 976 \\
% C052: identity_qasper_recomp_10
C052 & Q & RC/.1 & Id & $P_g$ & $-0.426$ & $[-0.456,-0.397]$ & 0.0148 & 976 \\
% C053: identity_qasper_recomp_5
C053 & Q & RC/.05 & Id & $P_g$ & $-0.425$ & $[-0.454,-0.396]$ & 0.0148 & 976 \\
% C054: identity_qasper_lingua_50
C054 & Q & LL/.5 & Id & $P_g$ & $-0.345$ & $[-0.381,-0.309]$ & 0.0148 & 976 \\
% C055: identity_qasper_lingua_25
C055 & Q & LL/.25 & Id & $P_g$ & $-0.372$ & $[-0.404,-0.338]$ & 0.0148 & 976 \\
% C056: identity_qasper_lingua_10
C056 & Q & LL/.1 & Id & $P_g$ & $-0.444$ & $[-0.471,-0.416]$ & 0.0148 & 976 \\
% C057: identity_qasper_lingua_5
C057 & Q & LL/.05 & Id & $P_g$ & $-0.466$ & $[-0.493,-0.439]$ & 0.0148 & 976 \\
% C058: identity_asqa_rerank_25_recall
C058 & A & RR/.25 & Id & $R_g$ & $+0.074$ & $[+0.039,+0.107]$ & 0.0148 & 944 \\
% C059: ecr_extract_asqa_50
C059 & A & ECR/.5 & X/.5 & $P_g$ & $+0.010$ & $[-0.018,+0.038]$ & 1.0000 & 944 \\
% C060: ecr_extract_asqa_25
C060 & A & ECR/.25 & X/.25 & $P_g$ & $-0.006$ & $[-0.036,+0.022]$ & 1.0000 & 944 \\
\bottomrule
\end{tabular}
\caption{Planned comparisons 31--60 of 148. $\Delta$: left-minus-right mean difference; $p_H$: Holm-adjusted $p$-value. Setting and metric labels follow Appendix~\ref{app:statistics}.}
\label{tab:paired-02}
\end{table*}

% AUTO-GENERATED by scripts/build_step2_tables.py. DO NOT EDIT.
\begin{table*}[t]
\centering
\small
\setlength{\tabcolsep}{3pt}
\begin{tabular}{ll ll c r c r r}
\toprule
ID & Data & Left & Right & Endpoint & $\Delta$ & 95\% CI & $p_H$ & $n$ \\
\midrule
% C061: ecr_extract_asqa_10
C061 & A & ECR/.1 & X/.1 & $P_g$ & $-0.004$ & $[-0.029,+0.021]$ & 1.0000 & 944 \\
% C062: ecr_extract_asqa_5
C062 & A & ECR/.05 & X/.05 & $P_g$ & $-0.027$ & $[-0.049,-0.005]$ & 0.4158 & 944 \\
% C063: ecr_extract_qasper_50
C063 & Q & ECR/.5 & X/.5 & $P_g$ & $+0.063$ & $[+0.033,+0.092]$ & 0.0148 & 976 \\
% C064: ecr_extract_qasper_25
C064 & Q & ECR/.25 & X/.25 & $P_g$ & $+0.030$ & $[+0.002,+0.059]$ & 1.0000 & 976 \\
% C065: ecr_extract_qasper_10
C065 & Q & ECR/.1 & X/.1 & $P_g$ & $+0.025$ & $[-0.001,+0.051]$ & 1.0000 & 976 \\
% C066: ecr_extract_qasper_5
C066 & Q & ECR/.05 & X/.05 & $P_g$ & $+0.018$ & $[-0.009,+0.046]$ & 1.0000 & 976 \\
% C067: lingua_precision_asqa_50_25
C067 & A & LL/.25 & LL/.5 & $P_g$ & $-0.053$ & $[-0.076,-0.029]$ & 0.0148 & 944 \\
% C068: lingua_precision_asqa_25_10
C068 & A & LL/.1 & LL/.25 & $P_g$ & $-0.051$ & $[-0.073,-0.030]$ & 0.0148 & 944 \\
% C069: lingua_precision_asqa_10_5
C069 & A & LL/.05 & LL/.1 & $P_g$ & $-0.018$ & $[-0.036,-0.001]$ & 1.0000 & 944 \\
% C070: lingua_precision_qasper_50_25
C070 & Q & LL/.25 & LL/.5 & $P_g$ & $-0.027$ & $[-0.058,+0.004]$ & 1.0000 & 976 \\
% C071: lingua_precision_qasper_25_10
C071 & Q & LL/.1 & LL/.25 & $P_g$ & $-0.072$ & $[-0.096,-0.048]$ & 0.0148 & 976 \\
% C072: lingua_precision_qasper_10_5
C072 & Q & LL/.05 & LL/.1 & $P_g$ & $-0.022$ & $[-0.041,-0.005]$ & 0.3836 & 976 \\
% C073: rerank_peak_asqa_25_50
C073 & A & RR/.25 & RR/.5 & $P_g$ & $+0.086$ & $[+0.053,+0.117]$ & 0.0148 & 944 \\
% C074: rerank_peak_asqa_25_10
C074 & A & RR/.25 & RR/.1 & $P_g$ & $+0.317$ & $[+0.280,+0.354]$ & 0.0148 & 944 \\
% C075: rerank_peak_asqa_25_5
C075 & A & RR/.25 & RR/.05 & $P_g$ & $+0.451$ & $[+0.415,+0.487]$ & 0.0148 & 944 \\
% C076: quality_identity_asqa_rerank_topk_50
C076 & A & RR/.5 & Id & $\mathrm{EM}$ & $-0.039$ & $[-0.055,-0.023]$ & 0.0148 & 944 \\
% C077: quality_identity_asqa_rerank_topk_25
C077 & A & RR/.25 & Id & $\mathrm{EM}$ & $-0.091$ & $[-0.108,-0.074]$ & 0.0148 & 944 \\
% C078: quality_identity_asqa_rerank_topk_10
C078 & A & RR/.1 & Id & $\mathrm{EM}$ & $-0.153$ & $[-0.172,-0.134]$ & 0.0148 & 944 \\
% C079: quality_identity_asqa_rerank_topk_5
C079 & A & RR/.05 & Id & $\mathrm{EM}$ & $-0.229$ & $[-0.249,-0.209]$ & 0.0148 & 944 \\
% C080: quality_identity_asqa_extractive_50
C080 & A & X/.5 & Id & $\mathrm{EM}$ & $-0.029$ & $[-0.044,-0.014]$ & 0.0148 & 944 \\
% C081: quality_identity_asqa_extractive_25
C081 & A & X/.25 & Id & $\mathrm{EM}$ & $-0.082$ & $[-0.100,-0.064]$ & 0.0148 & 944 \\
% C082: quality_identity_asqa_extractive_10
C082 & A & X/.1 & Id & $\mathrm{EM}$ & $-0.151$ & $[-0.170,-0.132]$ & 0.0148 & 944 \\
% C083: quality_identity_asqa_extractive_5
C083 & A & X/.05 & Id & $\mathrm{EM}$ & $-0.199$ & $[-0.219,-0.179]$ & 0.0148 & 944 \\
% C084: quality_identity_asqa_ecr_50
C084 & A & ECR/.5 & Id & $\mathrm{EM}$ & $-0.029$ & $[-0.045,-0.014]$ & 0.0216 & 944 \\
% C085: quality_identity_asqa_ecr_25
C085 & A & ECR/.25 & Id & $\mathrm{EM}$ & $-0.075$ & $[-0.093,-0.057]$ & 0.0148 & 944 \\
% C086: quality_identity_asqa_ecr_10
C086 & A & ECR/.1 & Id & $\mathrm{EM}$ & $-0.138$ & $[-0.157,-0.120]$ & 0.0148 & 944 \\
% C087: quality_identity_asqa_ecr_5
C087 & A & ECR/.05 & Id & $\mathrm{EM}$ & $-0.202$ & $[-0.222,-0.183]$ & 0.0148 & 944 \\
% C088: quality_identity_asqa_recomp_50
C088 & A & RC/.5 & Id & $\mathrm{EM}$ & $-0.054$ & $[-0.070,-0.039]$ & 0.0148 & 944 \\
% C089: quality_identity_asqa_recomp_25
C089 & A & RC/.25 & Id & $\mathrm{EM}$ & $-0.054$ & $[-0.069,-0.039]$ & 0.0148 & 944 \\
% C090: quality_identity_asqa_recomp_10
C090 & A & RC/.1 & Id & $\mathrm{EM}$ & $-0.055$ & $[-0.070,-0.040]$ & 0.0148 & 944 \\
\bottomrule
\end{tabular}
\caption{Planned comparisons 61--90 of 148. $\Delta$: left-minus-right mean difference; $p_H$: Holm-adjusted $p$-value. Setting and metric labels follow Appendix~\ref{app:statistics}.}
\label{tab:paired-03}
\end{table*}

% AUTO-GENERATED by scripts/build_step2_tables.py. DO NOT EDIT.
\begin{table*}[t]
\centering
\small
\setlength{\tabcolsep}{3pt}
\begin{tabular}{ll ll c r c r r}
\toprule
ID & Data & Left & Right & Endpoint & $\Delta$ & 95\% CI & $p_H$ & $n$ \\
\midrule
% C091: quality_identity_asqa_recomp_5
C091 & A & RC/.05 & Id & $\mathrm{EM}$ & $-0.067$ & $[-0.084,-0.051]$ & 0.0148 & 944 \\
% C092: quality_identity_asqa_lingua_50
C092 & A & LL/.5 & Id & $\mathrm{EM}$ & $-0.041$ & $[-0.055,-0.026]$ & 0.0148 & 944 \\
% C093: quality_identity_asqa_lingua_25
C093 & A & LL/.25 & Id & $\mathrm{EM}$ & $-0.101$ & $[-0.118,-0.083]$ & 0.0148 & 944 \\
% C094: quality_identity_asqa_lingua_10
C094 & A & LL/.1 & Id & $\mathrm{EM}$ & $-0.184$ & $[-0.203,-0.165]$ & 0.0148 & 944 \\
% C095: quality_identity_asqa_lingua_5
C095 & A & LL/.05 & Id & $\mathrm{EM}$ & $-0.227$ & $[-0.248,-0.207]$ & 0.0148 & 944 \\
% C096: quality_adjacent_asqa_rerank_topk_50_25
C096 & A & RR/.25 & RR/.5 & $\mathrm{EM}$ & $-0.052$ & $[-0.064,-0.040]$ & 0.0148 & 944 \\
% C097: quality_adjacent_asqa_rerank_topk_25_10
C097 & A & RR/.1 & RR/.25 & $\mathrm{EM}$ & $-0.062$ & $[-0.077,-0.046]$ & 0.0148 & 944 \\
% C098: quality_adjacent_asqa_rerank_topk_10_5
C098 & A & RR/.05 & RR/.1 & $\mathrm{EM}$ & $-0.076$ & $[-0.092,-0.060]$ & 0.0148 & 944 \\
% C099: quality_adjacent_asqa_extractive_50_25
C099 & A & X/.25 & X/.5 & $\mathrm{EM}$ & $-0.053$ & $[-0.068,-0.037]$ & 0.0148 & 944 \\
% C100: quality_adjacent_asqa_extractive_25_10
C100 & A & X/.1 & X/.25 & $\mathrm{EM}$ & $-0.069$ & $[-0.084,-0.055]$ & 0.0148 & 944 \\
% C101: quality_adjacent_asqa_extractive_10_5
C101 & A & X/.05 & X/.1 & $\mathrm{EM}$ & $-0.048$ & $[-0.062,-0.034]$ & 0.0148 & 944 \\
% C102: quality_adjacent_asqa_ecr_50_25
C102 & A & ECR/.25 & ECR/.5 & $\mathrm{EM}$ & $-0.046$ & $[-0.060,-0.032]$ & 0.0148 & 944 \\
% C103: quality_adjacent_asqa_ecr_25_10
C103 & A & ECR/.1 & ECR/.25 & $\mathrm{EM}$ & $-0.064$ & $[-0.079,-0.048]$ & 0.0148 & 944 \\
% C104: quality_adjacent_asqa_ecr_10_5
C104 & A & ECR/.05 & ECR/.1 & $\mathrm{EM}$ & $-0.064$ & $[-0.079,-0.049]$ & 0.0148 & 944 \\
% C105: quality_adjacent_asqa_recomp_50_25
C105 & A & RC/.25 & RC/.5 & $\mathrm{EM}$ & $+0.000$ & $[-0.002,+0.002]$ & 1.0000 & 944 \\
% C106: quality_adjacent_asqa_recomp_25_10
C106 & A & RC/.1 & RC/.25 & $\mathrm{EM}$ & $-0.001$ & $[-0.006,+0.004]$ & 1.0000 & 944 \\
% C107: quality_adjacent_asqa_recomp_10_5
C107 & A & RC/.05 & RC/.1 & $\mathrm{EM}$ & $-0.012$ & $[-0.022,-0.003]$ & 0.3030 & 944 \\
% C108: quality_adjacent_asqa_lingua_50_25
C108 & A & LL/.25 & LL/.5 & $\mathrm{EM}$ & $-0.060$ & $[-0.076,-0.045]$ & 0.0148 & 944 \\
% C109: quality_adjacent_asqa_lingua_25_10
C109 & A & LL/.1 & LL/.25 & $\mathrm{EM}$ & $-0.083$ & $[-0.100,-0.067]$ & 0.0148 & 944 \\
% C110: quality_adjacent_asqa_lingua_10_5
C110 & A & LL/.05 & LL/.1 & $\mathrm{EM}$ & $-0.043$ & $[-0.057,-0.030]$ & 0.0148 & 944 \\
% C111: evidence_ecr_extract_qasper_50
C111 & Q & ECR/.5 & X/.5 & $\mathrm{EF1}$ & $-0.0007$ & $[-0.0011,-0.0002]$ & 0.0374 & 976 \\
% C112: evidence_ecr_extract_qasper_25
C112 & Q & ECR/.25 & X/.25 & $\mathrm{EF1}$ & $-0.0018$ & $[-0.0025,-0.0012]$ & 0.0148 & 976 \\
% C113: evidence_ecr_extract_qasper_10
C113 & Q & ECR/.1 & X/.1 & $\mathrm{EF1}$ & $-0.0022$ & $[-0.0035,-0.0010]$ & 0.0148 & 976 \\
% C114: evidence_ecr_extract_qasper_5
C114 & Q & ECR/.05 & X/.05 & $\mathrm{EF1}$ & $-0.0030$ & $[-0.0046,-0.0012]$ & 0.0185 & 976 \\
% C115: evidence_adjacent_qasper_extractive_50_25
C115 & Q & X/.25 & X/.5 & $\mathrm{EF1}$ & $+0.0199$ & $[+0.0170,+0.0227]$ & 0.0148 & 976 \\
% C116: evidence_adjacent_qasper_extractive_25_10
C116 & Q & X/.1 & X/.25 & $\mathrm{EF1}$ & $+0.0389$ & $[+0.0343,+0.0435]$ & 0.0148 & 976 \\
% C117: evidence_adjacent_qasper_extractive_10_5
C117 & Q & X/.05 & X/.1 & $\mathrm{EF1}$ & $+0.0289$ & $[+0.0231,+0.0346]$ & 0.0148 & 976 \\
% C118: evidence_adjacent_qasper_ecr_50_25
C118 & Q & ECR/.25 & ECR/.5 & $\mathrm{EF1}$ & $+0.0187$ & $[+0.0160,+0.0214]$ & 0.0148 & 976 \\
% C119: evidence_adjacent_qasper_ecr_25_10
C119 & Q & ECR/.1 & ECR/.25 & $\mathrm{EF1}$ & $+0.0385$ & $[+0.0339,+0.0430]$ & 0.0148 & 976 \\
% C120: evidence_adjacent_qasper_ecr_10_5
C120 & Q & ECR/.05 & ECR/.1 & $\mathrm{EF1}$ & $+0.0281$ & $[+0.0222,+0.0340]$ & 0.0148 & 976 \\
\bottomrule
\end{tabular}
\caption{Planned comparisons 91--120 of 148. $\Delta$: left-minus-right mean difference; $p_H$: Holm-adjusted $p$-value. Setting and metric labels follow Appendix~\ref{app:statistics}.}
\label{tab:paired-04}
\end{table*}

% AUTO-GENERATED by scripts/build_step2_tables.py. DO NOT EDIT.
\begin{table*}[t]
\centering
\small
\setlength{\tabcolsep}{3pt}
\begin{tabular}{ll ll c r c r r}
\toprule
ID & Data & Left & Right & Endpoint & $\Delta$ & 95\% CI & $p_H$ & $n$ \\
\midrule
% C121: evidence_adjacent_qasper_rerank_topk_50_25
C121 & Q & RR/.25 & RR/.5 & $\mathrm{EF1}$ & $-0.0162$ & $[-0.0225,-0.0097]$ & 0.0148 & 976 \\
% C122: evidence_adjacent_qasper_rerank_topk_25_10
C122 & Q & RR/.1 & RR/.25 & $\mathrm{EF1}$ & $-0.0194$ & $[-0.0275,-0.0112]$ & 0.0148 & 976 \\
% C123: evidence_adjacent_qasper_rerank_topk_10_5
C123 & Q & RR/.05 & RR/.1 & $\mathrm{EF1}$ & $-0.0084$ & $[-0.0171,+0.0008]$ & 1.0000 & 976 \\
% C124: recovery_adjacent_asqa_lingua_50_25
C124 & A & LL/.25 & LL/.5 & $\mathrm{Rec}$ & $-0.156$ & $[-0.194,-0.117]$ & 0.0148 & 944 \\
% C125: recovery_adjacent_asqa_lingua_25_10
C125 & A & LL/.1 & LL/.25 & $\mathrm{Rec}$ & $-0.153$ & $[-0.192,-0.114]$ & 0.0148 & 944 \\
% C126: recovery_adjacent_asqa_lingua_10_5
C126 & A & LL/.05 & LL/.1 & $\mathrm{Rec}$ & $-0.050$ & $[-0.086,-0.015]$ & 0.2418 & 944 \\
% C127: verification_modes_asqa_identity
C127 & A & Id & Id & $U_s-U_m$ & $-0.002$ & $[-0.007,+0.002]$ & 1.0000 & 944 \\
% C128: verification_modes_asqa_rerank_topk
C128 & A & RR/.25 & RR/.25 & $U_s-U_m$ & $-0.003$ & $[-0.008,+0.002]$ & 1.0000 & 944 \\
% C129: verification_modes_asqa_extractive
C129 & A & X/.25 & X/.25 & $U_s-U_m$ & $+0.000$ & $[-0.002,+0.001]$ & 1.0000 & 944 \\
% C130: verification_modes_asqa_ecr
C130 & A & ECR/.25 & ECR/.25 & $U_s-U_m$ & $+0.009$ & $[+0.002,+0.016]$ & 0.3045 & 944 \\
% C131: verification_modes_qasper_identity
C131 & Q & Id & Id & $U_s-U_m$ & $-0.001$ & $[-0.005,+0.003]$ & 1.0000 & 976 \\
% C132: verification_modes_qasper_rerank_topk
C132 & Q & RR/.25 & RR/.25 & $U_s-U_m$ & $-0.002$ & $[-0.006,+0.001]$ & 1.0000 & 976 \\
% C133: verification_modes_qasper_extractive
C133 & Q & X/.25 & X/.25 & $U_s-U_m$ & $-0.008$ & $[-0.014,-0.003]$ & 0.0627 & 976 \\
% C134: verification_modes_qasper_ecr
C134 & Q & ECR/.25 & ECR/.25 & $U_s-U_m$ & $-0.001$ & $[-0.007,+0.004]$ & 1.0000 & 976 \\
% C135: tau_gap_asqa_recomp_0.3
C135 & A & RC/.25 & RC/.25/.3 & $P_e-P_g$ & $+0.714$ & $[+0.685,+0.743]$ & 0.0148 & 944 \\
% C136: tau_gap_asqa_recomp_0.4
C136 & A & RC/.25 & RC/.25/.4 & $P_e-P_g$ & $+0.730$ & $[+0.701,+0.758]$ & 0.0148 & 944 \\
% C137: tau_gap_asqa_recomp_0.6
C137 & A & RC/.25 & RC/.25/.6 & $P_e-P_g$ & $+0.753$ & $[+0.726,+0.780]$ & 0.0148 & 944 \\
% C138: tau_gap_asqa_recomp_0.7
C138 & A & RC/.25 & RC/.25/.7 & $P_e-P_g$ & $+0.766$ & $[+0.739,+0.792]$ & 0.0148 & 944 \\
% C139: tau_gap_asqa_recomp_0.8
C139 & A & RC/.25 & RC/.25/.8 & $P_e-P_g$ & $+0.779$ & $[+0.752,+0.805]$ & 0.0148 & 944 \\
% C140: tau_gap_asqa_lingua_0.3
C140 & A & LL/.25 & LL/.25/.3 & $P_e-P_g$ & $+0.372$ & $[+0.338,+0.406]$ & 0.0148 & 944 \\
% C141: tau_gap_asqa_lingua_0.4
C141 & A & LL/.25 & LL/.25/.4 & $P_e-P_g$ & $+0.400$ & $[+0.367,+0.433]$ & 0.0148 & 944 \\
% C142: tau_gap_asqa_lingua_0.6
C142 & A & LL/.25 & LL/.25/.6 & $P_e-P_g$ & $+0.449$ & $[+0.415,+0.481]$ & 0.0148 & 944 \\
% C143: tau_gap_asqa_lingua_0.7
C143 & A & LL/.25 & LL/.25/.7 & $P_e-P_g$ & $+0.474$ & $[+0.441,+0.506]$ & 0.0148 & 944 \\
% C144: tau_gap_asqa_lingua_0.8
C144 & A & LL/.25 & LL/.25/.8 & $P_e-P_g$ & $+0.488$ & $[+0.455,+0.520]$ & 0.0148 & 944 \\
% C145: recovered_verification_asqa_recomp_25
C145 & A & RC/.25 & RC/.25 & $U_r-U_m$ & $+0.712$ & $[+0.683,+0.739]$ & 0.0148 & 944 \\
% C146: recovered_verification_asqa_lingua_25
C146 & A & LL/.25 & LL/.25 & $U_r-U_m$ & $+0.394$ & $[+0.362,+0.426]$ & 0.0148 & 944 \\
% C147: recovered_verification_qasper_recomp_25
C147 & Q & RC/.25 & RC/.25 & $U_r-U_m$ & $+0.450$ & $[+0.417,+0.483]$ & 0.0148 & 974 \\
% C148: recovered_verification_qasper_lingua_25
C148 & Q & LL/.25 & LL/.25 & $U_r-U_m$ & $+0.253$ & $[+0.225,+0.282]$ & 0.0148 & 976 \\
\bottomrule
\end{tabular}
\caption{Planned comparisons 121--148 of 148. $\Delta$: left-minus-right mean difference; $p_H$: Holm-adjusted $p$-value. Setting and metric labels follow Appendix~\ref{app:statistics}.}
\label{tab:paired-05}
\end{table*}

\clearpage
\twocolumn
\section{Second-Evaluator Sensitivity Audit}
\label{app:true-audit}

We sample 100 questions per dataset from sorted common-cohort IDs using a fresh Python seed-13 random generator; QASPER spans 84 papers.
Full-context identity and nominal-0.25 extraction, RECOMP, and LLMLingua-2 contribute 800 unchanged saved answers, with 1,627 claims and 1,890 citation markers.
The prior DeBERTa values are the original scores for these same sampled answers.

\paragraph{Model and decoding.}
We use \texttt{google/t5\_xxl\_true\_nli\_mixture} \citep{honovich2022true}, revision \texttt{aa6cfe1dd4}, with full revision and checkpoint hashes in the manifest.
This released mixture uses SNLI, MNLI, FEVER, SciTail, PAWS, and VitaminC \citep{bowman2015snli,williams2018mnli,thorne2018fever,khot2018scitail,zhang2019paws,schuster2021vitaminc}, rather than the original TRUE paper's ANLI recipe.
MNLI and FEVER overlap DeBERTa's training mixture.
Inference uses Transformers 4.57.6, the slow T5 tokenizer, BF16, batch size four, and one NVIDIA B200.
The prompt places compressed text or source evidence after \texttt{premise:} and the unit or claim after \texttt{hypothesis:}.

We use an AutoAIS-style exact-one decision \citep{bohnet2022attributed,gao2023alce}: generate at most ten tokens deterministically, strip surrounding whitespace, and count only \texttt{1} as entailment.
Every other string scores zero; these are binary decisions, not calibrated probabilities.
Unlike our rule, reference AutoAIS/ALCE code compares decoded strings without additional whitespace normalization.
We retain normalized nonbinary strings, but not every raw output, so normalization's effect cannot be quantified from the cache.

\paragraph{Evidence mappings.}
\emph{Fixed recovery} preserves declared provenance and DeBERTa-recovered source IDs while TRUE scores support.
\emph{TRUE recovery} reevaluates all top-10 lexical candidates for undeclared units, including DeBERTa-rejected candidates, then scores with TRUE.
Both retain the corrected precision gate.
Fixed recovery retains selection dependence; TRUE recovery reuses one judge.
Neither removes bounded-search errors or supplies human calibration.

\paragraph{Recorded amendments.}
A strict-decoding pilot stopped on a nonbinary output; a second stopped on an overlong hypothesis.
An input-only scan found five undeclared units exceeding 2,048 tokens (maximum hypothesis plus prompt: 2,988), motivating a uniform 4,096-token cap that preserves complete hypotheses and truncates only premises.
A later batch-eight run exhausted GPU memory.
All interrupted attempts and plans are preserved but excluded; the final batch-four run recomputes every input in a fresh cache.
These amendments preceded inspection of any aggregate TRUE comparison report and left the sample, thresholds, and 24 comparisons unchanged.
The cap exceeds original TRUE's 2,048 and DeBERTa's 512 tokens, so cross-evaluator differences do not isolate model family.

\paragraph{Inference and diagnostics.}
The separate 24-test family uses pointwise 95\% bootstrap intervals and two-sided randomization tests, each with 10,000 replicates and seed 13, question resampling on ASQA and paper clustering on QASPER, with question-weighted means and Holm adjustment.
The original 148-test analysis is unchanged.
Tables~\ref{tab:true-audit-precision}--\ref{tab:true-audit-contrasts} report precision means and all planned contrasts.
All 16 emitted--grounded gaps remain positive after adjustment; the eight cross-evaluator precision differences remain unresolved, which does not establish equivalence.
ECR and other budgets are outside this audit.

The run scored 6,015 unique pairs in 419 seconds of model inference (430 seconds including bookkeeping; setup and excluded attempts are additional).
Four inputs required truncation, with maximum pre-truncation length 47,440 tokens.
After normalization, 122 pairs (2.03\%) produced nonbinary strings, including 103 empty strings; assigning zero does not establish reliable non-entailment.
These are pair counts, not fractions of questions affected; the full histogram accompanies the cache.
Recovery can disagree substantially: QASPER LLMLingua-2 has 50.3\% agreement over 1,000 candidate decisions and mean recovery rates of 0.84 for DeBERTa versus 0.38 for TRUE.
These descriptive results support sensitivity analysis, not accuracy against human judgments.

% AUTO-GENERATED by scripts/build_true_audit_tables.py. DO NOT EDIT.
\begin{table*}[t]
\centering
\small
\setlength{\tabcolsep}{3pt}
\begin{tabular}{llcccccc}
\toprule
& & \multicolumn{2}{c}{Prior DeBERTa} & \multicolumn{2}{c}{TRUE fixed} & \multicolumn{2}{c}{TRUE recovery} \\
\cmidrule(lr){3-4}\cmidrule(lr){5-6}\cmidrule(lr){7-8}
Data & Method & $P_e$ & $P_g$ & $P_e$ & $P_g$ & $P_e$ & $P_g$ \\
\midrule
A & Identity & 0.467 & 0.467 & 0.563 & 0.563 & 0.563 & 0.563 \\
A & Extractive & 0.402 & 0.402 & 0.497 & 0.497 & 0.497 & 0.497 \\
A & RECOMP & 0.790 & 0.090 & 0.780 & 0.130 & 0.780 & 0.140 \\
A & LLMLingua-2 & 0.480 & 0.070 & 0.415 & 0.023 & 0.415 & 0.010 \\
\midrule
Q & Identity & 0.487 & 0.487 & 0.498 & 0.498 & 0.498 & 0.498 \\
Q & Extractive & 0.461 & 0.461 & 0.391 & 0.391 & 0.391 & 0.391 \\
Q & RECOMP & 0.550 & 0.020 & 0.440 & 0.010 & 0.440 & 0.030 \\
Q & LLMLingua-2 & 0.490 & 0.118 & 0.387 & 0.076 & 0.387 & 0.019 \\
\bottomrule
\end{tabular}
\caption{Precision on the same sampled saved answers: 100 ASQA (A) and 100 QASPER (Q) questions (84 QASPER papers). $P_e/P_g$: emitted/grounded precision. Mappings and evaluation details follow Appendix~\ref{app:true-audit}; superscripts give nonmissing counts and dashes undefined values.}
\label{tab:true-audit-precision}
\end{table*}

% AUTO-GENERATED by scripts/build_true_audit_tables.py. DO NOT EDIT.
\begin{table*}[t]
\centering
\small
\setlength{\tabcolsep}{3pt}
\begin{tabular}{lllccrcrrr}
\toprule
ID & Data & Method & View & Endpoint & $\Delta$ & 95\% CI & $p_H$ & $n$ & $k$ \\
\midrule
T01 & A & RC & $F$ & $P_e-P_g$ & $+0.650$ & $[+0.550, +0.750]$ & 0.0024 & 100 & 100 \\
T02 & A & RC & $F$ & $R_e-R_g$ & $+0.640$ & $[+0.540, +0.735]$ & 0.0024 & 100 & 100 \\
T03 & A & LL & $F$ & $P_e-P_g$ & $+0.392$ & $[+0.290, +0.495]$ & 0.0024 & 100 & 100 \\
T04 & A & LL & $F$ & $R_e-R_g$ & $+0.367$ & $[+0.273, +0.463]$ & 0.0024 & 100 & 100 \\
T05 & Q & RC & $F$ & $P_e-P_g$ & $+0.430$ & $[+0.326, +0.530]$ & 0.0024 & 100 & 84 \\
T06 & Q & RC & $F$ & $R_e-R_g$ & $+0.425$ & $[+0.320, +0.525]$ & 0.0024 & 100 & 84 \\
T07 & Q & LL & $F$ & $P_e-P_g$ & $+0.311$ & $[+0.234, +0.390]$ & 0.0024 & 100 & 84 \\
T08 & Q & LL & $F$ & $R_e-R_g$ & $+0.233$ & $[+0.176, +0.293]$ & 0.0024 & 100 & 84 \\
\midrule
T09 & A & RC & $T$ & $P_e-P_g$ & $+0.640$ & $[+0.540, +0.740]$ & 0.0024 & 100 & 100 \\
T10 & A & RC & $T$ & $R_e-R_g$ & $+0.637$ & $[+0.540, +0.730]$ & 0.0024 & 100 & 100 \\
T11 & A & LL & $T$ & $P_e-P_g$ & $+0.405$ & $[+0.312, +0.502]$ & 0.0024 & 100 & 100 \\
T12 & A & LL & $T$ & $R_e-R_g$ & $+0.375$ & $[+0.287, +0.468]$ & 0.0024 & 100 & 100 \\
T13 & Q & RC & $T$ & $P_e-P_g$ & $+0.410$ & $[+0.309, +0.505]$ & 0.0024 & 100 & 84 \\
T14 & Q & RC & $T$ & $R_e-R_g$ & $+0.405$ & $[+0.303, +0.505]$ & 0.0024 & 100 & 84 \\
T15 & Q & LL & $T$ & $P_e-P_g$ & $+0.368$ & $[+0.286, +0.452]$ & 0.0024 & 100 & 84 \\
T16 & Q & LL & $T$ & $R_e-R_g$ & $+0.263$ & $[+0.204, +0.325]$ & 0.0024 & 100 & 84 \\
\midrule
T17 & A & Id & $F-D$ & $P_g$ & $+0.096$ & $[+0.011, +0.185]$ & 0.2376 & 100 & 100 \\
T18 & A & X & $F-D$ & $P_g$ & $+0.095$ & $[+0.024, +0.168]$ & 0.0880 & 100 & 100 \\
T19 & A & RC & $F-D$ & $P_g$ & $+0.040$ & $[-0.010, +0.100]$ & 0.8810 & 100 & 100 \\
T20 & A & LL & $F-D$ & $P_g$ & $-0.047$ & $[-0.097, +0.000]$ & 0.6299 & 100 & 100 \\
T21 & Q & Id & $F-D$ & $P_g$ & $+0.010$ & $[-0.074, +0.098]$ & 0.9887 & 100 & 84 \\
T22 & Q & X & $F-D$ & $P_g$ & $-0.070$ & $[-0.134, -0.008]$ & 0.1827 & 100 & 84 \\
T23 & Q & RC & $F-D$ & $P_g$ & $-0.010$ & $[-0.025, +0.000]$ & 0.9887 & 100 & 84 \\
T24 & Q & LL & $F-D$ & $P_g$ & $-0.042$ & $[-0.095, +0.011]$ & 0.6299 & 100 & 84 \\
\bottomrule
\end{tabular}
\caption{All 24 planned TRUE robustness contrasts, separate from the original 148-test analysis. $D/F/T$: prior DeBERTa/TRUE fixed/TRUE recovery; $n/k$: questions/clusters. $\Delta$: left-minus-right mean difference; $p_H$: Holm-adjusted within this 24-test family. 10,000 replicates, seed 13. Other notation follows Appendix~\ref{app:statistics}.}
\label{tab:true-audit-contrasts}
\end{table*}

\end{document}